\documentclass{bmvc2k}
\usepackage{framed,multirow}

\usepackage{amssymb}
\usepackage{latexsym}
\usepackage{graphicx}
\usepackage[font=scriptsize]{caption}
\usepackage{graphicx}
\usepackage{amsmath}
\usepackage{amssymb}
\usepackage{booktabs}
\usepackage{arydshln}
\usepackage{algorithm}
\usepackage{algpseudocode}
\usepackage{multirow}

\usepackage{microtype}
\usepackage{graphicx}
\usepackage{subfigure}
\usepackage{booktabs} % for professional tables

\usepackage{hyperref}

\usepackage{amsmath}
\usepackage{amssymb}
\usepackage{mathtools}
\usepackage{amsthm}

\usepackage{graphicx}

\usepackage[utf8]{inputenc} % allow utf-8 input
\usepackage[T1]{fontenc}    % use 8-bit T1 fonts
\usepackage{url}            % simple URL typesetting
\usepackage{booktabs}       % professional-quality tables
\usepackage{amsfonts}       % blackboard math symbols
\usepackage{nicefrac}       % compact symbols for 1/2, etc.
\usepackage{microtype}      % microtypography
\usepackage{xcolor}         % colors

\usepackage{times}
\usepackage{epsfig}
\usepackage{dirtytalk}
\usepackage{placeins}
\usepackage{float}
\usepackage{enumerate}
\usepackage{wrapfig}

\usepackage{algorithm}
\usepackage{algpseudocode}

\usepackage{multirow, booktabs}

\usepackage{pifont}% http://ctan.org/pkg/pifont
\usepackage[capitalize,noabbrev]{cleveref}

\title{Attentive Contractive Flow with Lipschitz Constrained Self-Attention }
\addauthor{Avideep Mukherjee}{avideep@cse.iitk.ac.in}{1}
\addauthor{Badri N. Patro}{badri.patro@kuleuven.be}{2}
\addauthor{Vinay P. Namboodiri}{vpn22@bath.ac.uk}{3}

\addinstitution{
 Indian Institute of Technology Kanpur\\
 Kanpur, 208016\\
 Uttar Pradesh, India
}
\addinstitution{
 KU Leuven\\
 Oude Markt 13, 3000\\
 Leuven, Belgium
}
\addinstitution{
 University of Bath\\
 Claverton Down, Bath\\
 BA2 7AY, United Kingdom
}

\runninghead{Mukherjee, Patro, Namboodiri}{Attentive Contractive Flow}

\begin{document}

\maketitle

\begin{abstract}
Normalizing flows provide an elegant method for obtaining tractable density estimates from distributions using invertible transformations. The main challenge is improving the models' expressivity while keeping the invertibility constraints intact. We propose to do so via the incorporation of localized self-attention. However, conventional self-attention mechanisms do not satisfy the requirements to obtain invertible flows and cannot be naively incorporated into normalizing flows. To address this, we introduce a novel approach called Attentive Contractive Flow (ACF) which utilizes a special category of flow-based generative models - contractive flows. We demonstrate that ACF can be introduced into various state-of-the-art flow models in a plug-and-play manner. This is demonstrated to improve the representation power of these models (improving on the bits per dim metric) and result in significantly faster convergence in training them. Qualitative results, including interpolations between test images, demonstrate that samples are more realistic and capture local correlations in the data well. We evaluate the results further by performing perturbation analysis using AWGN demonstrating that ACF models (especially the dot-product variant) show better and more consistent resilience to additive noise.
\end{abstract}

%-------------------------------------------------------------------------
% \vspace{-2em}
\section{Introduction}
\label{sec:intro}
While deep generative models based on generative adversarial networks (GANs) and variational autoencoders (VAEs) produce state-of-the-art results showing impressive results on megapixel images, they do not have the ability to obtain exact likelihood estimates. To address this need, flow-based approaches such as real NVP  \cite{dinh2016density} and invertible residual networks  \cite{behrmann2019invertible} have been proposed. However, flow-based models are still limited in their modelling capabilities as compared to GANs and VAEs. This paper focuses on improving the modelling capability of these models through the incorporation of self-attention in them. For this purpose, a sub-category of flow-based generative models called contractive flows \cite{rezende2015variational} is leveraged. 

Incorporating attention in normalizing flows while maintaining the constraints of invertibility and tractable computation of the log-determinant of the Jacobian is a challenge. We propose a solution using contractive flows, called Attentive Contractive Flows (ACF), which improves modelling capability, convergence during training, and resilience to input noise.

Recent progress in generative modeling has been made through GANs such as StyleGAN2 \cite{Karras_2020} and hierarchical variational autoencoders like NVAE \cite{vahdat2020nvae} and Flow++ \cite{ho2019flow++}. While Flow++ incorporates self-attention to a slice of the image in the coupling layer, incorporating attention to the entire image space in normalizing flow-based models has not been proposed yet.

Self-attention \cite{vaswani2017attention} can achieve localized importance in the image and latent space, balancing the ability to model inter-dependent features with computational and statistical efficiency. During density estimation of high dimensional data, attention helps to obtain information about key positions in the image that are representative of a sample. Self-attention has been successfully incorporated in GANs\cite{zhang2019self} and VAEs\cite{lin2019improving}, but incorporating it in Normalizing Flows in a generic sense is challenging due to the models' different transformation functions.

A main challenge we need to solve in order to achieve this task is to examine the  invertibility of the self-attention module. In this paper, we show that while the general self-attention module \cite{vaswani2017attention} is not uniquely invertible, it can be made Lipschitz-continuous by replacing the dot-product operation with an $L_2$ norm \cite{kim2020lipschitz} or by normalizing the whole function by a certain scalar quantity \cite{dasoulas2021lipschitz}. Therefore, self-attention can be made into a contraction and can be incorporated in three different contractive flows: iResNet  \cite{behrmann2019invertible}, Residual Flows  \cite{chen2019residual} and iDenseNets  \cite{perugachi2021invertible} each one being an improvement over the previous one. We show that the performance of all three contractive flows gets better, respectively, with Self Attention. A contractive flow uses Banach's Fixed Point Theorem to guarantee exact iterative inverses of arbitrarily complex neural networks as long as the neural network function remains a contraction. We use the variant of the self-attention mechanism inside a contractive neural network to provide a perfectly law-abiding flow-based generative model. Through this model, we are also able to obtain improved expressive capability for obtaining flows and show significantly improved performance with fewer steps as compared to other state-of-the-art NF models. Our main contributions can be summarized as follows:
\begin{itemize}
    \item It is evident that self-attention plays a major role in the improved performance of a number of deep learning architectures. The analysis of the same for the flow-based generative approach has not been so well considered.
    \vspace{-1.5em}
    \item In this work, we show that naive self-attention is mathematically not tractable. We show this empirically through comparisons. We need to consider the Lipschitz norm while including self-attention in normalizing flows. We demonstrate two variants of the same by incorporating either the $L_2$ norm or using the Lipschitz normalization and obtain attentive contractive flows.
    \vspace{-1.5em}
    \item These have been shown to perform competitively to state-of-the-art normalizing flow methods and can be considered a complementary way for improving them.
\end{itemize}

 \begin{figure}[t]
     \small
     \centering
   \includegraphics[width=0.7\textwidth]{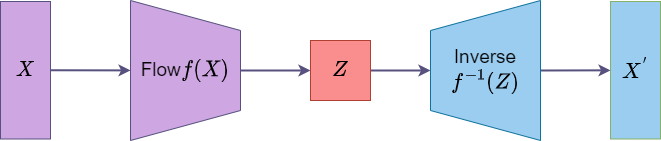}
   \vspace{1em}
      \caption{Attentive Contractive Flow (ACF) based generative model.}
      \label{fig:diagram}
 \end{figure}
% \vspace{-1.em}

\section{Preliminaries}\label{background}

\subsection{Flow-Based Generative Models}
Normalizing Flows are a class of generative models where an \textit{‘simple’ parameterizable base distribution} is transformed into a \textit{more complex approximation} for the posterior distribution  \cite{rezende2015variational}). This transformation is achieved by passing the base distribution through a series of invertible and bijective mappings. Let $\mathbf{z} \in \mathbb{R}^D$ and $\mathbf{y = f(z)}$ Let $\mathbf{z} \sim q(\mathbf{z})$, be an simple base distribution. The
change of variables theorem expresses a relation between the probability density functions $p_Y(\mathbf y)$ and $q(\mathbf z)$:
\begin{equation}\small
    p_Y(\mathbf{y}) = q(\mathbf{z}) \left|
    \mathrm{det} \frac{
      \partial f^{-1}
    }{
      \partial \mathbf{y}\
    }
  \right|
  = q(\mathbf{z}) \left|
    \mathrm{det} \frac{
      \partial f
    }{
      \partial \mathbf{z}\
    }
  \right| ^{-1}.
\end{equation}
If we apply a series of such mappings $f_k$, $k \in {1,\dots, K}$ with $K \in \mathbb{N}_+$, we obtain a \textbf{normalizing flow}. The log probability of the final distribution can thus be obtained by:
\begin{equation}\small\label{eq_2}
\mbox{\fontsize{10.5pt}{9pt}\selectfont\( %
       \log p_Y(\mathbf{y}) = \log p_Y(\mathbf{z}_K) = \log q(\mathbf{z}_0) - \sum_{k=1}^{K} \log
    \left| \mathrm{det} \frac{\partial{f_k}}{\partial{\mathbf{z}_{k-1}}}
    \right|.
\)} %
\end{equation}
From Equations 1 \& 2, it is clearly observed that every normalizing flow architecture must satisfy two conditions. First, the transformation function should be invertible. Secondly, the log-determinant of the Jacobian should be tractable.

\subsection{Contractive Flows}
\label{contrflow}
% \subsubsection{iResNet \& Residual Flows}
Residual Flows  \cite{papamakarios2019normalizing} are a class of invertible functions of the form:
\begin{equation}\small
\label{restrans}
     \mathbf{z' = f(z) = z + g_\phi(z)}
\end{equation}
where $g_\phi : \mathbb{R}^D \mapsto \mathbb{R}^D$ is a neural network with parameters $\phi$.  \cite{behrmann2019invertible} proposed the transformation of Equation \ref{restrans} as an invertible residual network or iResNet. Residual transformations like this can be made invertible with certain constraints on $g_\phi$.  \cite{behrmann2019invertible} show that a residual transformation is guaranteed to be invertible if the function $g_\phi$ is a \textit{contraction}. A  \textit{contraction} is a special case of a Lipschitz continuous function. A function $F : \mathbb{R}^D \mapsto \mathbb{R}^D$ is said to be $K$-Lipschitz continuous when for a given distance measure $\delta$, there exists a constant $K$ such that for two inputs $\mathbf{x_1}$ and $\mathbf{x_2}$ we have: $\delta(F(\mathbf{x_1}),F(\mathbf{x_2})) \leq K\delta(\mathbf{x_1},\mathbf{x_2}).$
% \vspace{-0.5em}\begin{equation}\small
%     \delta(F(\mathbf{x_1}),F(\mathbf{x_2})) \leq K\delta(\mathbf{x_1},\mathbf{x_2}).
% \end{equation}
The smallest such $K$ is called the Lipschitz constant of $F$ or $\text{Lip}(F)$. If $\text{Lip}(F) \leq 1$, then $F$ is said to be a \textit{contraction}. Let us consider the following equation involving the contraction in Equation \ref{restrans}.
\begin{equation}\small
\label{inv}
    \mathbf{F(\hat{z}) = z' - g_\phi(\hat{z})} .
\end{equation}
Since $\mathbf{g_\phi}$ is contractive, $\mathbf{F}$ is also contractive with the same Lipschitz constant. Therefore, from \textit{Banach's Fixed Point Theorem}  \cite{rudin2006real}, it is ensured that there exists a unique $\mathbf{z_*}$ such that $\mathbf{z* = z' - g_\phi(z_*)}$ which can be rearranged to $\mathbf{z' = f(z_*)}$. Hence it follows that $\mathbf{f}$ is invertible  \cite{papamakarios2019normalizing}. In fact, the inversion algorithm is iteratively designed from Equation \ref{inv} as follows:
\begin{equation}\small
    z_{k+1} = z' - g_\phi(z_k)\quad \text{for}\quad k \geq 0.
\end{equation}
Banach's fixed-point theorem guarantees the convergence of the recursive algorithm at an exponential rate to $\mathbf{z_* = f^{-1}(z')}$ for any arbitrary initialization of $z_0$ (usually it is preferred to have $z_0 = z'$). It is clearly observed that the composition of $K$ such residual transformations also preserves the contractive properties with the Lipschitz constant being $\prod_{i=1}^K L_K$, where $L_K$ is the respective Lipschitz constant of $F_K$. However, there are two major challenges to building residual flows. First off, the design of the neural network function is restricted to being Lipschitz continuous, that too contraction, which limits the flexibility of the network. Secondly, the calculation of the log-determinant of the Jacobian of such a transformation cannot be efficiently computed except for automatic differentiation, which takes $\mathcal{O}(\mathcal{D}^3)$ time. However, the log-determinant can be approximated using the results of  \cite{hall2015lie} and  \cite{withers2010log} and re-written as a power series of the trace of the Lipschitz network $\mathbf{g_\phi}$:

% \vspace{-0.5em}\begin{equation}\small\label{eq_7}
% \mbox{\fontsize{11pt}{11pt}\selectfont\( %
%     \begin{split}
%     \log|\det J_{f_\phi}(\mathbf{z})| &= \log|\det \left(\mathbf{I} + J_{g_\phi}(\mathbf{z})\right)|\\
%                  & = \sum_{k=1}^\infty\frac{(-1)^{k+1}}{k}\text{Tr}\left\{J^k_{g_\phi}(\mathbf{z})\right\}
% \end{split}
% \)} %
% \end{equation}
% \vspace{-0.5em}\begin{equation}\small\label{eq_7}
%       \log|\det J_{f_\phi}(\mathbf{z})| &= \log|\det \left(\mathbf{I} + J_{g_\phi}(\mathbf{z})\right)| = \sum_{k=1}^\infty\frac{(-1)^{k+1}}{k}\text{Tr}\left\{J^k_{g_\phi}(\mathbf{z})\right\}
%   \end{equation}
\vspace{-0.5em}\begin{equation}\small\label{eq_7}
  \log|\det J_{f_\phi}(\mathbf{z})| = \log|\det \left(\mathbf{I} + J_{g_\phi}(\mathbf{z})\right)| + \sum_{k=1}^\infty\frac{(-1)^{k+1}}{k}\text{Tr}\left\{J^k_{g_\phi}(\mathbf{z})\right\}
\end{equation}

where $J^k_{g_\phi}(\mathbf{z})$ is the $k$-th power of the Jacobian of $g_\phi$ at $\mathbf{z}$.The trace can be estimated using the Hutchinson trace estimator  \cite{hutchinson1989stochastic}. \textbf{Residual Flows}  \cite{chen2019residual} improved this method where the power series can be finitely approximated using the unbiased Russian-roulette estimator. This results in a lower requirement of computation of the power series than in the case of iResNet. Also, they introduce the LipSwish activation function to avoid derivative saturation.

% \vspace{-1.5em}
% \subsubsection{Invertible DenseNets} 
Invertible DenseNets  \cite{perugachi2021invertible} use a DenseBlock as a residual layer. A DenseBlock in iDenseNet is slightly different than a standard DenseBlock and is defined as $F: \mathbb{R}^d \mapsto \mathbb{R}^d$ with $F(x) = x + g(x)$ where $g$ is comprised of dense layers $\{h_i\}_{i=i}^{n+1}$. $h_{n+1}$ is a $1 \times 1$ convolution to match the dimension of the output size $\mathbb{R}^d$. Each $h_i$ has two concatenated parts, the input and the transformed input: 
\begin{align}\small
    h_i(x) &= \begin{bmatrix}
           x \\
           \phi(W_i(x)) \\
         \end{bmatrix}
  \end{align}
 where $W_i$ is convolutional matrix and $\phi$ is a non-linearity with Lip($\phi$) $\leq$ 1 such as ReLU, ELU, LipSwish  \cite{chen2019residual} or CLipSwish  \cite{perugachi2021invertible}.

\subsection{Self-Attention with Lipschitz Continuity}
Lipschitz Continuity is important in neural networks to stabilize training and mitigate problems like gradient explosion. For residual flows to become invertible, they must be Lipschitz Continuous, requiring all transformation function modules to be so. Two methods of ensuring Lipschitz Continuity in self-attention modules are explored.
\subsubsection{$L_2$ Self Attention}
\label{l2-mha}
\cite{kim2020lipschitz} proved that the standard self-attention module introduced by \cite{vaswani2017attention}, or the \textit{dot-product self-attention}, as they called it (since it involved computing a dot-product to generate the attention maps) is not Lipschitz continuous, and as a result, it is unsuitable for use in methods such as residual flows. Furthermore, they have proposed a Lipschitz Continuous variant of the self-attention function called the $L_2$ Self Attention. The $L_2$ self-attention function on an image $X \in \mathbb{R}^{N\times D}$ (where $N$ is the product of the height and width of the image and $D$ is the number of channels) replaces the dot-product operation performed in dot-product self-attention module by:
\vspace{-1em}\begin{equation}\small
\label{$L_2$-prop}
    P_{i,j} \propto \text{exp}(L_{i,j}) = \text{exp}\left(-\frac{||x_i^TW^Q -x_j^TW^K||^2_2}{\sqrt{D/H}}\right).
\end{equation}
% \vspace{-1em}
Here $W^Q = W^K \in \mathbb{R}^{D \times \Bar{D}}$ and $H$ is the number of heads in the multi-headed self-attention function. So, the full $L_2$ Multi-headed Self-Attention function is given by:
% \vspace{-0.5em}
\begin{equation}\small
\label{L_2}
    F(X) = [f^1(X)W^{V,1}, \cdots, f^H(X)W^{V,H}]W^O
\end{equation}
where, $f^h(X) = P^hXA_h$, $W^{V,h} \in \mathbb{R}^{D \times D/H}$, $W^O \in \mathbb{R}^{D \times D}$, $P^h$ is defined as in Equation \ref{$L_2$-prop} with $W^{Q,h} = W^{K,h} \in \mathbb{R}^{D \times D/H}$ and $A_h = W^{Q,h}{W^{Q,h}}^T/\sqrt{D/H} \in \mathbb{R}^{D \times D}$.

\subsubsection{Lipschitz Normalization}
\label{lipnorm}
% \vspace{-0.5em}
\cite{dasoulas2021lipschitz} introduced a normalization scheme on the dot-product self-attention itself and which makes the function Lipschitz continuous. Let $\Tilde{g}(X): \mathbb{R}^{d \times n} \mapsto \mathbb{R}^{m \times n}$ be the score function of an attention model that takes an input matrix and returns scores for each output vector $i \in {1, \cdots, m}$ and each input vector $j \in {1, \cdots, n}$. \cite{dasoulas2021lipschitz} normalized $\Tilde{g}$ by some scalar function $c: \mathbb{R}^{d \times n} \mapsto \mathbb{R}_+$ and proved that under certain assumptions, the normalized score function $g(X) = \Tilde{g}(X)/c(X)$, is Lipschitz Continuous with the following bound:
% \vspace{-0.5em}
\begin{equation}\small
  \small
    L_F(Att) \leq e^{\alpha} \sqrt{\frac{m}{n}} + \alpha\sqrt{8}
\end{equation}
where $\alpha$ controls the scale of all the scores.

\section{Method: Contractive Flows with Self-Attention}
% \vspace{-1em}
Before defining the transformation function with self-attention, we need to make sure that the two conditions of the normalizing flow function are satisfied, i.e., the function is invertible, and the log-determinant of the Jacobian is tractable. As shown in Figure \ref{fig:diagram}, on a complex image space $X$, and given a simple base distribution $z$, the model tries to learn an invertible bijective function $f$ that maps $X$ to $Z$. In our case, $f$ is a contraction making the whole system a contractive flow. So, in order to have attention incorporated in contractive flows, the attention module needs to be inserted in between the convolutional layers in the function $\mathbf{g}_\phi$ of Equation \ref{restrans}. This would require the attention function to be a Lipschitz continuous function. Since the attention function involves computing a dot-product between the query and key matrices, it will be referred to as \textit{dot-product self-attention}.
Even if the dot-product self-attention module can be inverted, the log-determinant of the Jacobian is hard to compute.
\begin{figure}[t]
  \centering
\includegraphics[width=0.46\textwidth]{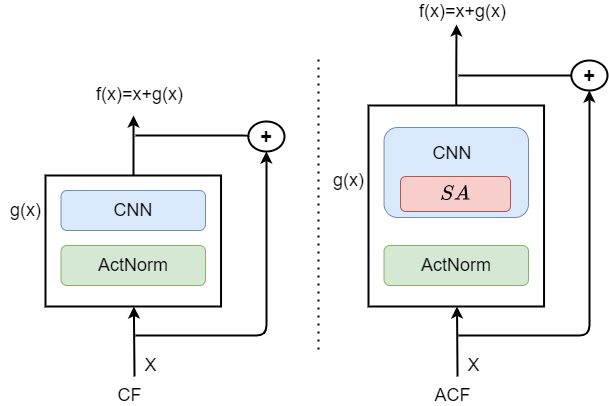}
   \vspace{1em}
   \caption{Model diagram comparing one step of the transformation function of a generic contractive flow (CF) (left) and a contractive flow with Self Attention (ACF)(right). SA stands for Self Attention.}
   \label{fig:main}
   \vspace{-1em}
\end{figure}
 A trick to compute the non-tractable log-determinant of the Jacobian is to use the results of  \cite{withers2010log} that reduces computing the determinant to computing the trace. The result shows that for any non-singular matrix $A \in \mathbb{R}^{d\times d}$, $\ln(\det A) = \text{tr}(\ln A)$
% \begin{equation*}
%     \ln(\det A) = \text{tr}(\ln A)
% \end{equation*}
where $\ln$ is a matrix logarithm, $\text{tr}$ is the trace of a matrix.
However, this result requires the determinant be a positive quantity. It is proven that the Lipschitz-constrained perturbations of the form $x + g(x)$ yield positive Jacobian determinants  \cite{behrmann2019invertible}. Therefore, for a function $F(x) = x + g(x)$, we have, $ |\det J_F(x)| = \det J_F(x).$
% \vspace{-0.5em}\begin{equation}\small
%     |\det J_F(x)| = \det J_F(x).
% \end{equation}

 Here comes the role of contractive flows, which defines transformation functions as $F(x) = x + g(x)$. But, as explained in Section \ref{contrflow}, in order to make the transformation function invertible, $g$ should not only be Lipschitz continuous but also needs to be a \textit{contraction}, i.e., $\text{Lip}(g) < 1$. Since the dot-product self-attention is not Lipschitz continuous, as stated earlier, we use other variations of self-attention as discussed in Section \ref{l2-mha} and \ref{lipnorm}. Equation \ref{$L_2$-prop} can be rewritten using matrix operations for efficient computation as:
 \vspace{-0.5em}\begin{equation}\small\label{eq_13}
   \small
   P = \mathbf{S}\left(-\frac{\|XW^Q\|_{r}^2\mathbf{1}^T - 2XW^Q{(XW^K)}^T + \mathbf{1}\|XW^K\|_{r}^{2T}}{\sqrt{D/H}}\right)
   \end{equation}
  %  \vspace{-0.5em}
 where \textbf{S} is the softmax function and $||A||^2_r$ indicates the squared $L_2$ norm to each row of $A$, so if $A \in \mathbb{R}^{m \times n}$, then $||A||^2_r \in \mathbb{R}^m$.
 As discussed in Section \ref{l2-mha} this formulation of self-attention is proven to be Lipschitz continuous and with the following bound on $\text{Lip}_2(F)$:
 \vspace{-0.5em}\begin{equation}\small\label{eq_14}
   \small
   \text{Lip}_2(F) \leq \frac{\sqrt{N}}{\sqrt{D}}(4\phi^{-1}(N-1)+1)\|W^Q\|_2\|W^V\|_2\|W^O\|_2
   \end{equation}
   
 where $\phi(x) = x\exp(x+1)$ is an invertible univariate function on $x > 0$ and $N$ is the input size. Also, $\phi^{-1}(N-1) = W_0(\frac{N}{e})$ where $W_0$ is the Lambert W-function  \cite{kim2020lipschitz}.

\begin{algorithm}[t]
  
  % \vspace{-1.0em}

  \caption{Pseudo Code for a forward pass of an ACF with $L_2$ Self Attention \cite{kim2020lipschitz}. SN stands for Spectral Normalization as in \cite{behrmann2019invertible}}\label{algo:forward}

  \begin{algorithmic}[1]
  \Require network $f$, residual block $g$, number of power series terms $n$, $W^Q$: the query convolution, $W^V,W^O$ : the value and out convolution respectively, $H$: the number of heads in the multi-headed self-attention block.
  \Require $X \in \mathbb{R}^{N\times D}$ (where $N$ is the product of the height and width of the image and $D$ is the number of channels)
  
  \For{each residual block} 
    
    \State Lip constraint: $\hat{W_j}$ := $SN(W_j, X)$ for Layer $W_j$ 

    \State $P = \mathbf{S}\left(-\frac{\|XW^Q\|_{r}^2\mathbf{1}^T - 2XW^Q{(XW^K)}^T + \mathbf{1}\|XW^K\|_{r}^{2T}}{\sqrt{D/H}}\right)$
    \State $A := W^Q(W^Q)^T/\sqrt{D}$
    \State $F := P \times X \times A \times W^L$ as in eq:~\ref{L_2}
    \State $\text{Lip}_2(F) := \frac{\sqrt{N}}{\sqrt{D}}(4\phi^{-1}(N-1)+1)||W^Q||_2||W^V||_2W^O||_2$
    \State $\hat{W_{j+1}} := \gamma\frac{F}{\text{Lip}_2(F)} + X$: the final attention output as mentioned in eq:~\ref{eq:out}
    \State Draw $v$ from $\mathcal{N}(0,\mathbf{I})$
    \State $w^T = v^T$
    \State ln det := 0
    \For{k = 1 \textbf{to} n}
      \State $w^T:= {w^T}J_g$ (vector-Jacobian product)
      \State ln det:= ln det $+ (-1)^{k+1}{w^T}v/k$
    \EndFor
  \EndFor
\end{algorithmic}
\end{algorithm}
% \begin{algorithm}[t]
  % \vspace{-1em}
%   % \vspace{-1.0em}

%   \caption{Pseudo Code for a forward pass of an ACF with Lipschitz Normalization \cite{dasoulas2021lipschitz}. SN stands for Spectral Normalization as in \cite{behrmann2019invertible}}\label{algo:lipnorm}

%   \begin{algorithmic}[1]
%   \Require network $f$, residual block $g$, number of power series terms $n$, $W^Q$: the query convolution, $W^V,W^O$ : the value and out convolution respectively, $H$: the number of heads in the multi-headed self-attention block.
%   \Require $X \in \mathbb{R}^{N\times D}$ (where $N$ is the product of the height and width of the image and $D$ is the number of channels)
  
%   \For{each residual block} 
    
%     \State Lip constraint: $\hat{W_j}$ := $SN(W_j, X)$ for Layer $W_j$
%     \State $\Tilde{g}(X) = {W^Q}^TW^K$
%     \State $c(X) = \max\{uv,uw,vw\}$
%     \State $g(X) = \frac{g(X)}{c(X)}$ as in Section \ref{lipnorm}
%     \State $A = \text{softmax}(g(X))$
%     \State $F = {W^V}A^T$
%     \State $\text{Lip}_2(F) := e^{\sqrt{3}} \sqrt{\frac{m}{n}} + 2\sqrt{6}$
%     \State $\hat{W_{j+1}} := \gamma\frac{F}{\text{Lip}_2(F)} + X$: the final attention output as mentioned in eq:~\ref{eq:out}
%     \State Draw $v$ from $\mathcal{N}(0,\mathbf{I})$
%     \State $w^T = v^T$
%     \State ln det := 0
%     \For{k = 1 \textbf{to} n}
%       \State $w^T:= {w^T}J_g$ (vector-Jacobian product)
%       \State ln det:= ln det $+ (-1)^{k+1}{w^T}v/k$
%     \EndFor
%   \EndFor
% \end{algorithmic}

% % \vspace{-0.2em}

% \end{algorithm}
%\vspace{-1em}
Hence, in order to make $F$ a contraction, we divide $F$ by the upper bound of $\text{Lip}_2(F)$ to obtain contractive-$L_2$ Self-Attention. This function satisfies every property of being a part of the transformation function of the normalizing flow. Therefore, the final attention output is given by:
\vspace{-0.5em}\begin{equation}\small
    \text{out} = \gamma\frac{F}{\text{Lip}_2(F)} + X
    \label{eq:out}
\end{equation}
where $\gamma$ is a learnable scalar initialized to 0. $\gamma$ helps the network to first attend to the local features in the neighbourhood and then gradually learn to assign more weight to the non-local evidence  \cite{zhang2019self}. A detailed procedure of the forward pass of one step of the flow along with the  computation of the $L_2$ Self Attention is provided in Algorithm \ref{algo:forward}.

Alternatively, we also applied Lipschitz Normalization on dot-product self-attention as described in Section \ref{lipnorm}. In the context of residual flows, we use the transformer variant of Lipschitz normalization. In the case of transformers, the score function, as mentioned in Section \ref{lipnorm}, is given by 
\vspace{-0.5em}\begin{equation}\small
    g(X) = \frac{Q^TK}{\max\{uv,uw,vw\}}
\end{equation}
where $u = ||Q||_F, v = ||K^T||_{(\infty,2)}$ and $w = ||V^T||_{(\infty,2)}$. This score function is Lipschitz with the following bound:
\vspace{-0.5em}\begin{equation}\small
    \text{Lip}_2(F) := e^{\sqrt{3}} \sqrt{\frac{m}{n}} + 2\sqrt{6}.
\end{equation}
Hence, similar to Equation \ref{eq:out}, the score function is divided by the bound to transform the function from being Lipschitz continuous to a contraction. A detailed procedure of the forward pass of one step of the flow, along with the  computation of the Lipschitz norm on dot-product self-attention, is provided in the supplementary material.

\vspace{-0.2em}
\section{Experiments}
\vspace{-0.2em}
We evaluate the inclusion of Self Attention in three different contractive flows: invertible ResNets, Residual Flows, and invertible DenseNets. We experiment with the combination of Lipschitz Normalization \cite{dasoulas2021lipschitz} on dot-product self-attention and residual flows \cite{chen2019residual} and test it on benchmark datasets. We analyze the rate of convergence of the flows both with and without attention (Fig \ref{fig:bpd}), do a qualitative analysis that includes visualizations (Fig \ref{samples_comparison} and \ref{sample_comparison_lip}), and conduct ablation studies to validate the efficacy of the method (Fig \ref{fig:interpolations}). 
\begin{table}[]
\scriptsize
\centering
\begin{tabular}{@{}ccccc@{}}
\toprule
\textbf{Model} & \textbf{MNIST} & \multicolumn{1}{l}{\shortstack{ \textbf{CIFAR}\textbf{10}}} & \multicolumn{1}{l}{\shortstack{ \textbf{IMAGE}\textbf{NET32}}} & \multicolumn{1}{l}{\shortstack{ \textbf{IMAGE}\textbf{NET64}}} \\ \midrule
Real NVP\cite{dinh2016density} & 1.06 & 3.49 & 4.28 & 3.98 \\
Glow\cite{kingma2018glow} & 1.05 & 3.35 & 4.09 & 3.81 \\
FFJORD\cite{grathwohl2018ffjord} & 0.99 & 3.40 & - & - \\
VFlow\cite{chen2020vflow} & -  & \textbf{2.98} & 3.83 & 3.66 \\
ANF\cite{huang2020augmented} & 0.93 & 3.05 & 3.92 & 3.66  \\
DenseFlow\cite{grcic2021densely} &  - & 2.98 & \textbf{3.63} & \textbf{3.35} \\
Flow++\cite{ho2019flow++} & -  & 3.29 & $3.86^*$ & $3.69^*$ \\ \cmidrule(r){1-5}
iResNet\cite{behrmann2019invertible} (iR) & 1.05 & 3.45 & - & - \\
 iR + $L_2$SA (ACF) & \textbf{0.87} & 3.40 & - & - \\\cmidrule(r){1-5}
ResFlow\cite{chen2019residual} (RF) & 0.97 & 3.28 & 4.01 & 3.76 \\
RF + $L_2$SA (ACF) & 0.90 & 3.34 & 3.86 & 3.70 \\
RF + LipNorm (ACF) & 1.14 & 3.05 & - & - \\\cmidrule(r){1-5}
iDenseNet\cite{perugachi2021invertible}  (iD) & - & 3.25 & 3.98 & - \\
iD + $L_2$SA (ACF) & - & 3.14 & 3.75 & -\\\hline
\end{tabular}
\vspace{1.5em}
\caption{ Results [bits/dim] on standard benchmark datasets for density estimation. * are the results obtained through variational dequantization  \cite{ho2019flow++} which we do not compare against (following Residual Flow). }
\label{tab:comparison}
\vspace{-2.5em}
\end{table}
\vspace{-0.5em}
\subsection{Datasets}\label{datasets}
\vspace{-0.5em}

The efficacy of ACF was experimentally validated using datasets like MNIST\cite{lecun1998mnist}, CIFAR10  \cite{krizhevsky2014cifar}, ImageNet32 and ImageNet64  \cite{chrabaszcz2017downsampled}, but not CelebA HQ 256 due to constrained resources. Instead, experiments were conducted on a down-sampled version: CelebA-HQ64 \cite{liu2015faceattributes}. The standard train-test split was followed for each dataset. More detailed information about the datasets used are provided in the supplementary materials.
% \vspace{-0.5em}
\subsection{Density Estimation and Generative Modelling} 

\begin{figure*}[h]
  \small
  \centering

  \begin{tabular}[b]{ccc }
    \includegraphics[width=0.30\linewidth]{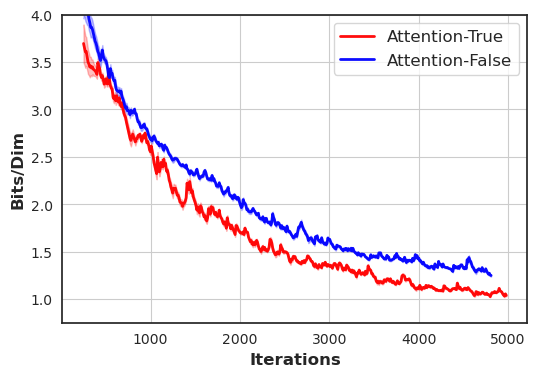}
    &\includegraphics[width=0.30\linewidth]{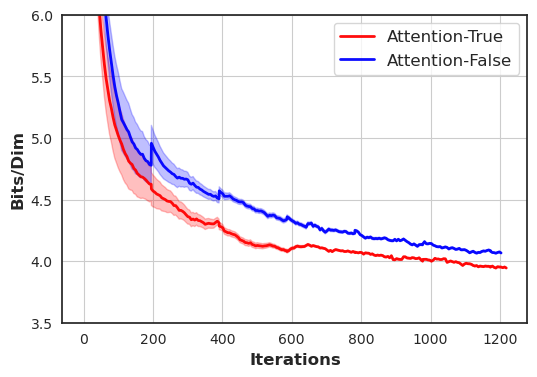}
    &\includegraphics[width=0.30\linewidth]{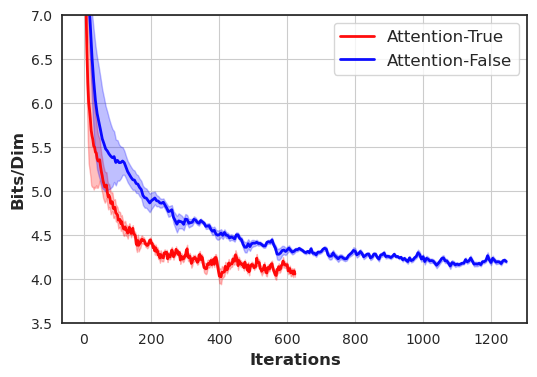}
    \end{tabular}
\vspace{0.5em}
\caption{Convergence plots on (left) iResNet on MNIST, (middle) Residual Flows on CIFAR10 and (right)  Invertible DenseNet on ImageNet32, in terms of train bits/dim across iterations. We observe that contractive flows with Self Attention converge faster (even with fewer steps of the flow) than their non-attention counterparts. All the experiments have been conducted with a random shuffling of the datasets, and the standard deviation is indicated with the shaded region after three independent trials.}
 \label{fig:bpd}
   \vspace{-0.2em}
\end{figure*}

We train ACF models (iResNet+L2SA, Residual Flow+L2SA, Residual Flow+ LipNorm, iDenseNet+L2SA) and their non-attentive counterparts on reported datasets. Multiple heads \cite{dosovitskiy2010image} are used in the L2 Self Attention block. The best results for MNIST and CIFAR10 datasets are achieved with H = 4 and H = 16, respectively (more in Section \ref{sec:heads}). ACF outperforms non-attentive contractive flows and other state-of-the-art models (Table \ref{tab:comparison}), while using fewer epochs (as compared to the compared methods) and achieving faster convergence rates (Fig \ref{fig:bpd}). We use the log determinant approximation (Section \ref{contrflow}) in all experiments and provide further model-specific details in supplementary material.
\vspace{-1.2em}
\subsection{Ablation Analysis: Choice of Self Attention Mechanism}
\label{sec:heads}

The authors experimented with multi-headed L2 Self Attention in the Residual Flow architecture, comparing it against single-headed self-attention. They used Residual Flow \cite{chen2019residual} and tested the model with H = 16 and H = 4 for multi-headed L2 Self Attention \cite{dosovitskiy2010image}. They reported the bits/dim performance in Table 2 for both MNIST and CIFAR10 datasets, finding that the multi-headed self-attention block performed slightly better than its single-headed counterpart. All other hyper-parameters were kept constant for all experiments.
% \vspace{1em}
\begin{table}
\vspace{1em}
\centering
\small
\begin{tabular}{ c|cc }  \hline
     \#heads & MNIST & CIFAR10\\
     \hline
     1 & 0.99 & 3.45 \\
     4 & \textbf{0.90} & 3.35 \\
     16 & 0.94 & \textbf{3.34}\\\hline
\end{tabular}
\vspace{1em}
\caption{Comparison of bits/dim results with Residual Flows + $L_2$ Multi-headed Self Attention varying on the number of heads tested on MNIST and CIFAR10 datasets.}
\label{tab:heads}
\vspace{-0.5em}
\end{table}

% \vspace{-1.2em}
\subsection{ Qualitative Results}
% \vspace{-0.5em}

\begin{figure}[h]
  \centering
  \subfigure{%
    \includegraphics[width=0.45\linewidth]{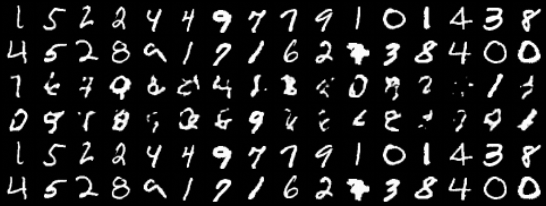}
  }
  % \hfill
  \subfigure{%
    \includegraphics[width=0.45\linewidth]{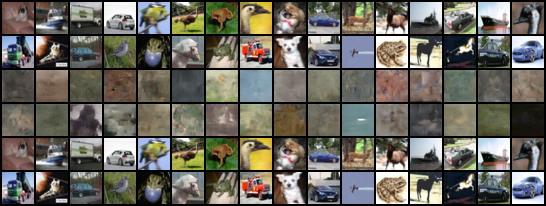}
  }
  \vspace{0.5em}
  \caption{The images in the top two, bottom two, and middle two rows are, respectively, the real, reconstructed and generated images. (a) MNSIT results from ACF (Residual Flows + LipNorm). (b) CIFAR10 results from ACF (Residual Flows + LipNorm).}
  \label{sample_comparison_lip}
\end{figure}

\begin{figure}[t]
  \centering
  \begin{tabular}{cc}
    \includegraphics[width=0.45\linewidth]{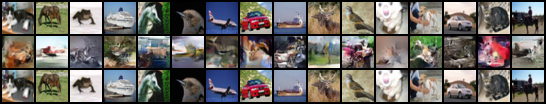} &
    \includegraphics[width=0.45\linewidth]{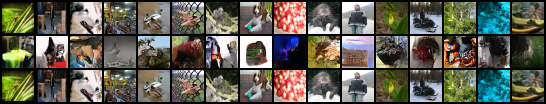}
  \end{tabular}
  \vspace{0.5em}
  \caption{The images in the top, bottom, and middle row are, respectively, the real, reconstructed, and generated images. (left) CIFAR10 results from ACF (Residual Flows + $L_2$SA). (right) ImageNet32 results from ACF (iDenseNet + $L_2$SA).}\label{samples_comparison}
   \vspace{-0.2em}
\end{figure}

\begin{figure}[t]
  \centering
  \subfigure{\includegraphics[width=0.43\linewidth]{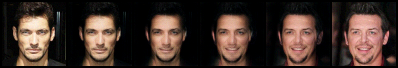}}
  %\hfill
  \subfigure{\includegraphics[width=0.43\linewidth]{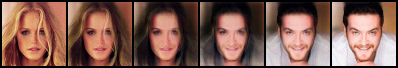}}
  
  \subfigure{\includegraphics[width=0.43\linewidth]{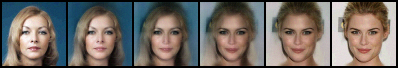}}
  %\hfill
  \subfigure{\includegraphics[width=0.43\linewidth]{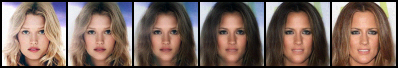}}
  \vspace{0.5em}
  \caption{Interpolation between CelebA images, from one face to another, using ACF(Residual Flow + $L_2$SA) and the increment rule provided \cite{sundararajan2017axiomatic}. }\label{fig:interpolations}
\end{figure}

Figures \ref{sample_comparison_lip} and \ref{samples_comparison} show qualitative samples and reconstructed images for ACF on MNIST, CIFAR10 and ImageNet32, demonstrating that it is capable of generating both exact reconstructions and realistic samples. Although there may be a discrepancy between sample quality and log-likelihood, ACF can still estimate the density of data better than other state-of-the-art generative models \cite{theis2015note}. Additionally, ACF has a better inductive bias than autoregressive flows due to being built upon residual blocks \cite{chen2019residual}. More samples are provided in the supplementary material.
\vspace{-0.5em}
\vspace{-0.5em}
% \vspace{-1em}
% We also train ACF(Residual Flow) on the CelebA dataset and perform interpolation between test images to provide a better visual intuition. We apply the transformation function (trained ACF) on the test images and obtain the corresponding samples in the latent space, i.e., the simple base distribution. We then perform linear interpolation between random pairs of images in the latent space, obtaining $N$ equally spaced interpolated latent samples per pair. We then apply the inverse of the transformation function on the interpolated latent samples to obtain their corresponding images. More formally, the input images, say $x_{1}$ and $x_{2}$ are passed through the forward transformation function of the model to generate the random variables from the simple distribution, say $z_{1}$ and $z_{2}$.
We also  perform interpolation between random pairs of CelebA test images (say $x_{1}$ and $x_{2}$) by applying the transformation function on them to obtain corresponding samples in the latent space (say $z_{1}$ and $z_{2}$). The interpolation is performed between random pairs of images to obtain equally spaced interpolated latent samples per pair. The intermediate random variables are generated by the incremental operation: $\delta= z_{1} + \frac{i}{N}\times(z_{2} - z_{1}) \quad \forall i \in \{0, 1, 2, \cdots, N+1\}$
% \vspace{-0.5em}\begin{equation}\small
% \label{ig}
%    \delta= z_{1} + \frac{i}{N}\times(z_{2} - z_{1}) \quad \forall i \in \{0, 1, 2, \cdots, N+1\}
% \end{equation}
. We follow the increment rule provided in the computation of integrated gradients  \cite{sundararajan2017axiomatic}. 
% The procedure to generate the interpolated reconstructions is described in Algorithm \ref{interpolate_algo}.
The results are depicted in Figure \ref{fig:interpolations}. We observe that ACF is able to interpolate effectively, providing smooth transitions between pairs of diverse and unseen faces. The algorithm for interpolation and more results are provided in the supplementary material.
% \begin{algorithm}[ht]
%   \DontPrintSemicolon
%      \KwInput{$f$: model, $nSteps$: interpolation steps)}%, $C_1$, $C_2$, $nSteps$: number of interpolation steps)}
%     \KwData{$x_{C_1}$ and $x_{C_2}$}
%     $z_{C_1}$ = $f(x_{C_1})$ and 
%     $z_{C_2}$ = $f(x_{C_2})$\\
%     \For{i $\in$ $\{0,1,2, \cdots, nSteps + 1\}$}
%       {
%           $\delta$ = $ z_{C_1} + \frac{i}{nSteps}\times(z_{C_2} - z_{C_1})$
%           \texttt{reconstructed\_image} = $f^{-1}(\delta)$
%       }
%   \caption{Interpolation Between Two Images}
%   \label{interpolate_algo}
%   \end{algorithm}
\vspace{-0.2em}
\vspace{-0.2em}
\subsection{Classification}
Following the underlying residual architecture in the attentive contractive flows, the models can be used in a discriminatory fashion and employed in tasks like image classification. 
% We have relevant experiments on the CIFAR10 dataset and chose ACF (Residual Flow) as a representative model for it. 
\begin{wraptable}{r}{5cm}
  % \vspace{-0.7em}
  \centering
  \scriptsize
  \begin{tabular}{ c|c}  \hline
      Model  & Accuracy (\%) \\ \hline
      Residual Flow \cite{chen2019residual}  & 91.78\\
      iDenseNet \cite{perugachi2021invertible}  &  92.40\\
      ACF (Ours) (iResNet + $L_2$SA) & \textbf{93.75}\\\hdashline
      ResNet v2-20 \cite{misra2019mish} & 92.2 \\
      ResNet9 + Mish \cite{misra2019mish} & 94.05\\
      ResNet + ELU \cite{shah2016deep} & 94.4 \\
      \hline\\
   
  \end{tabular}
  \caption{Comparison of accuracy on image classification of CIFAR10 dataset.}\label{tab:classification}\vspace{-1em}
  \end{wraptable}
  ACF (iResNet+$L_2$SA) achieves an accuracy of 93.75\% on test data of CIFAR10 in 200 fewer epochs of training compared to other flow-based methods. In Table \ref{tab:classification}, the model is compared with other contractive flows and also state-of-the-art discriminatory models that are based on residual architectures. Furthermore, since ACF preserves the contractive and residual structures as in  \cite{chen2019residual} and  \cite{perugachi2021invertible}, it is able to perform a discriminatory task while learning the underlying distribution of the data. Such hybrid modelling can be used in semi-supervised learning or anomaly detection.  \cite{chen2019residual} 
\vspace{-0.2em}
\section{Related Works}
\label{related}
\textbf{Attention in generative models}. Deep generative modelling aims to estimate the density or distribution of data to generate new samples. There are two approaches to estimation: implicit and explicit, leading to various techniques. GANs \cite{goodfellow2014generative} are the most famous among implicit density estimation models and use an adversarial game between a generator and a discriminator to generate new samples. The Self-Attention GAN \cite{zhang2019self}  uses a self-attention mechanism to improve results. Explicit density estimation models either approximate or tractably estimate the density. Apart from Restricted Boltzmann Machines \cite{smolensky1986information}, VAEs\cite{kingma2013auto} are a well-known example of approximate density estimation, optimizing the log-likelihood of the data by maximizing the evidence lower bound. Recent work suggests incorporating self-attention on the encoder feature space of VAEs to improve the approximation of the posterior.

\textbf{Normalizing Flows}. Normalizing flow models estimate the density of data and are part of the change of variable models category. Unlike GANs and VAEs, normalizing flows can obtain tractable density estimates. Popular normalizing flow models include NICE \cite{dinh2014nice}, Real NVP \cite{dinh2016density}, IAF \cite{kingma2016improved}, MAF \cite{papamakarios2017masked}, and Glow  \cite{kingma2018glow}. Recent methods like VFlow \cite{chen2020vflow} or ANF\cite{huang2020augmented} add additional dimensionalities to the data for better training and model expressivity. DenseFlow \cite[text]{grcic2021densely} proposes augmentation in the latent representation. Other approaches for generative modeling include NADE  \cite{uria2016neural}, MADE  \cite{kumar2016ask}, PixelCNN, PixelRNN  \cite{oord2016pixel} and WaveNet  \cite{oord2016wavenet}.The Flow++ \cite{ho2019flow++} model suggests implementing Self Attention in affine coupling layers of models such as real NVP \cite{dinh2016density}, but only provides attention to a slice of the image or feature space. This work proposes a method to incorporate Self Attention into normalizing flows in a way that attention is applied to the whole image and feature spaces, resulting in a greater improvement in performance.

\vspace{-0.2em}
\section{Conclusion and Future Direction}
\vspace{-0.2em} 
Our work presents a method to incorporate global self-attention into contractive flows for better modelling of complex image distributions. We show that the addition of $L_2$ self-attention or Lipschitz Normalization helps to attain state-of-the-art results and faster model convergence. Future research can explore how self-attention can improve specific model architectures and strengthen the representative power of normalizing flows.
\bibliography{egbib}
\end{document}

\maketitle

\section{Implementation and Hyperparameter Details}
\label{hyp}
The ACF models were experimented on the following datasets. Below are the details of the implementation and the hyper-parameters used. For all the experiments except ACF(iResNet) on CIFAR10, data parallelization has been performed. Tables 1, 2 and 3 provide details about the model architecture as well as the experimental environment it was trained in. The \textbf{Time} column in the tables refers to the time taken per epoch and is calculated as the product of the number of batches of the dataset and the time taken to train each batch.
\begin{table*}[ht]
  \centering
  
  \resizebox{\textwidth}{!}{
  
  \addtolength{\tabcolsep}{8.0pt}
  
  \begin{tabular}{ c | c | c | c | c | c | c | c }
      \toprule
      Dataset & Steps & Epochs & Batch Size & Params & \#GPU & GPU Type &  Time/Epoch (hr)\\
      \hline
      MNIST   & 10    & 120    & 16         & 3001090 &   1   &  TITAN X - 12GB & 6.2 \\
      CIFAR10 & 10    & 110    & 16         & 3001090 &   1   &  TITAN X - 12GB & 8.5 \\
      \bottomrule\\
  \end{tabular}
  
  }
  \caption{Implementation and Hyperparameter details on the respective datasets for ACF (iResNet + $L_2$SA)}
  \label{tab:hyperparameters}
\end{table*}
\vspace{-1.5em}
\begin{table*}[h]
  \centering
  \resizebox{\textwidth}{!}{

  \addtolength{\tabcolsep}{5.0pt}
  
  \begin{tabular}{ c | c | c | c | c | c | c | c }
      \toprule
      
      Dataset & Steps & Epochs & Batch Size & Params & \#GPU & GPU Type &Time/Epoch (hr)\\
      \hline
      MNIST   & 10    & 120    & 16         & 3597375 &  3 &  1080 Ti - 11GB & 4.7  \\
      CIFAR10 & 10    & 110    & 16         & 3597375 & 3 &  TITAN X - 12GB & 3.47 \\
      ImageNet32 & 16 & 1 & 16         & 5079333 &  2 & TITAN X - 12 GB & 164.6  \\
      ImageNet64 & 32 & 1 & 16         & 9031221 & 4  & Tesla V100-16GB   & 266.91\\
      5-bit CelebA - HQ &  8  & 100  & 64  & 3103389 & 4  & Tesla V100-16GB  &  0.58\\
      
  \bottomrule\\
  \end{tabular}}
  \caption{Implementation and Hyperparameter details on the respective datasets for ACF (Residual Flow + $L_2$SA)}\label{tab:hyperparameters}
\end{table*}
\vspace{-1em}
\begin{table*}[h]
  \centering
  \resizebox{\textwidth}{!}{

  \addtolength{\tabcolsep}{8.0pt}
  
  \begin{tabular}{ c|c|c|c|c|c|c|c }
      % \hline
      \toprule
      Dataset           & Steps & Epochs & Batch Size & Params & \#GPU & GPU Type &  Time/Epoch (hr) \\
      \hline
      CIFAR10           & 16    & 50    & 32         & 22577116 &  4   & Tesla V100-16GB  & 9.55\\
      ImageNet32        & 32    & 2      & 32         & 42719404 &  4   & Tesla V100-16GB  & 207.59\\
  \bottomrule\\
  \end{tabular}}
  \caption{Implementation and Hyperparameter details on the respective datasets for ACF (iDenseNet + $L_2$SA)}
  \label{tab:hyperparameters}
\end{table*}
\vspace{-1.5em}
\begin{table*}[h]
  \centering
  \resizebox{\textwidth}{!}{

  \addtolength{\tabcolsep}{8.0pt}
  
  \begin{tabular}{ c | c | c | c | c | c | c | c }
      % \hline
      \toprule
      Dataset & Steps & Epochs & Batch Size & Params & \#GPU & GPU Type &  Time/Epoch (hr) \\
      \hline
      MNIST & 10    & 120    & 16 & 2730495 &  3   & 1080 Ti-11GB  & 2.34\\
      CIFAR10 & 10    & 100      & 16 & 2730495 & 3   & 1080 Ti-11GB  & 4.89\\
  \bottomrule\\
  \end{tabular}}
  \caption{Implementation and Hyperparameter details on the respective datasets for ACF (Residual Flow + Lipschitz Normalization)}
  \label{tab:hyperparameters}
\end{table*}
\vspace{-1em}
\newpage
\section[short]{Algorithm for Contractive Flow with Lipschitz Normalized Self Attention}

\begin{algorithm}[h]
  
  % \vspace{-1.0em}

  \caption{Pseudo Code for a forward pass of an ACF with Lipschitz Normalization \cite{dasoulas2021lipschitz}. SN stands for Spectral Normalization as in \cite{behrmann2019invertible}}\label{algo:lipnorm}

  \begin{algorithmic}[1]
  \Require network $f$, residual block $g$, number of power series terms $n$, $W^Q$: the query convolution, $W^V,W^O$ : the value and out convolution respectively, $H$: the number of heads in the multi-headed self-attention block.
  \Require $X \in \mathbb{R}^{N\times D}$ (where $N$ is the product of the height and width of the image and $D$ is the number of channels)
  
  \For{each residual block} 
    
    \State Lip constraint: $\hat{W_j}$ := $SN(W_j, X)$ for Layer $W_j$
    \State $\Tilde{g}(X) = {W^Q}^TW^K$
    \State $c(X) = \max\{uv,uw,vw\}$
    \State $g(X) = \frac{g(X)}{c(X)}$ as in Section \ref{lipnorm}
    \State $A = \text{softmax}(g(X))$
    \State $F = {W^V}A^T$
    \State $\text{Lip}_2(F) := e^{\sqrt{3}} \sqrt{\frac{m}{n}} + 2\sqrt{6}$
    \State $\hat{W_{j+1}} := \gamma\frac{F}{\text{Lip}_2(F)} + X$: the final attention output as mentioned in eq:~\ref{eq:out}
    \State Draw $v$ from $\mathcal{N}(0,\mathbf{I})$
    \State $w^T = v^T$
    \State ln det := 0
    \For{k = 1 \textbf{to} n}
      \State $w^T:= {w^T}J_g$ (vector-Jacobian product)
      \State ln det:= ln det $+ (-1)^{k+1}{w^T}v/k$
    \EndFor
  \EndFor
\end{algorithmic}

% \vspace{-0.2em}
\end{algorithm}

\section{Datasets}
The CIFAR10 dataset consists of 60,000 $32 \times 32$ colour images belonging to 10 classes, with 6000 images per class. We train the models on 50,000 images and keep the rest 10,000 images to generate the test results. The MNIST dataset contains 70,000 $28 \times 28$ black and white images belonging to 10 classes, each class signifying a digit. We use 60,000 for the training of the models and the remaining 10,000 images for testing. The ImageNet32 and ImageNet64 datasets each comprise 1,281,167 training images from 1000 classes and 50,000 test images(50 images per class). The 5bit CelebA-HQ64 dataset contains 202,599 face images and is pre-processed as defined in  \cite{papamakarios2019normalizing}.

\section{Algorithm for Interpolation}
\begin{algorithm}[h]
  
  % \vspace{-1.0em}

  \caption{Interpolation Between Two Images}\label{interpolate_algo}

  \begin{algorithmic}[1]
  \Require $f$: model, $nSteps$: interpolation steps)
  \Require $C_1$, $C_2$, $nSteps$: number of interpolation steps
  \Require \textbf{Data:} $x_{C_1}$ and $x_{C_2}$

  \For{i $\in$ $\{0,1,2, \cdots, nSteps + 1\}$} 
    
    \State $\delta$ = $ z_{C_1} + \frac{i}{nSteps}\times(z_{C_2} - z_{C_1})$ 

    \State \texttt{reconstructed\_image} = $f^{-1}(\delta)$
  \EndFor
\end{algorithmic}

% \vspace{-0.2em}

\end{algorithm}
\section{Perturbation Analysis with Gaussian Noise}\label{subsec:noise}

\begin{figure*}[h]
    \centering
    \small
    
        \begin{tabular}[b]{ccc}
         (a) Input  & (b) $\epsilon (\sigma=0.1)$ DP-MHA $L_2$-MHA & (c)  $\epsilon(\sigma = 0.001)$ DP-MHA $L_2$-MHA \\ 
      \includegraphics[width=0.12\linewidth]{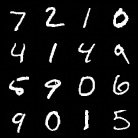}
      & \includegraphics[width=0.12\linewidth]{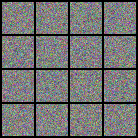}
       \includegraphics[width=0.12\linewidth]{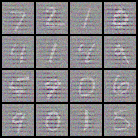}
     \includegraphics[width=0.12\linewidth]{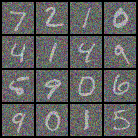}
    
     &\includegraphics[width=0.12\linewidth]{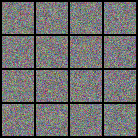}
    \includegraphics[width=0.12\linewidth]{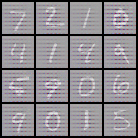}
    \includegraphics[width=0.12\linewidth]{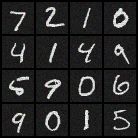}
    
       \end{tabular}
    \vspace{1em}
    \caption{Reconstructions of perturbed input images by a Contractive Flow with Dot and $L_2$ Self Attention. The perturbation is done by adding Gaussian noises ($\epsilon$) of different variances ($\sigma$), as shown in (b) and (c). ACF with $L_2$ Self Attention produces better reconstructions with a decrease in the intensity of noise. ACF with dot-product Self Attention fails to do so.}
    \label{recons_pert_$L_2$}
    % \vspace{-1.5em}
    \end{figure*}

\begin{figure*}[h]
  \vspace{-1em}
       \small
       \centering
       \begin{tabular}[b]{ c  c  c }
          %   \vspace{-3em} 
      %  (a) Analysis on bits/dim  & (b) Analysis on Logp(z)  & (c)  Analysis on Jacobian\\ 
      % \vspace{-2em} 
      \hspace{-3.3em}
      \includegraphics[width=0.3\linewidth]{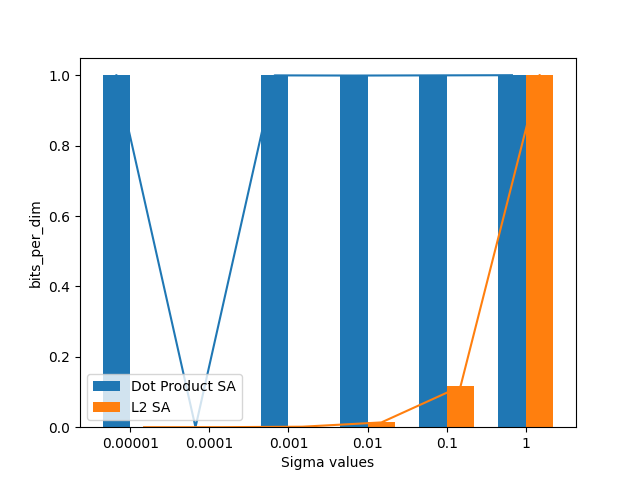}
      \hspace{-3.7em}
      
      &\includegraphics[width=0.3\linewidth]{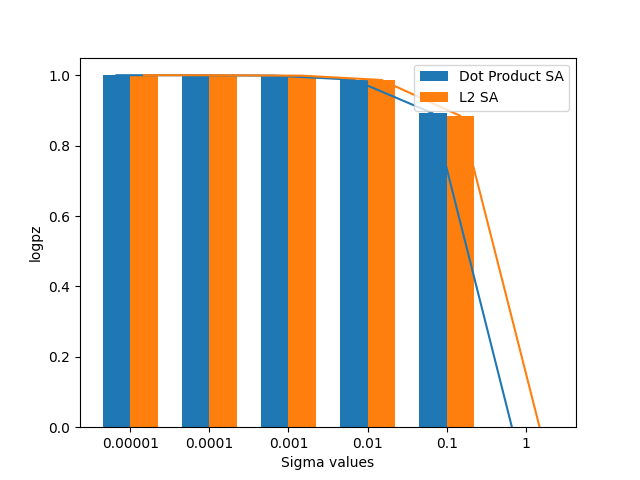}
      \hspace{-3.3em}
      &\includegraphics[width=0.3\linewidth]{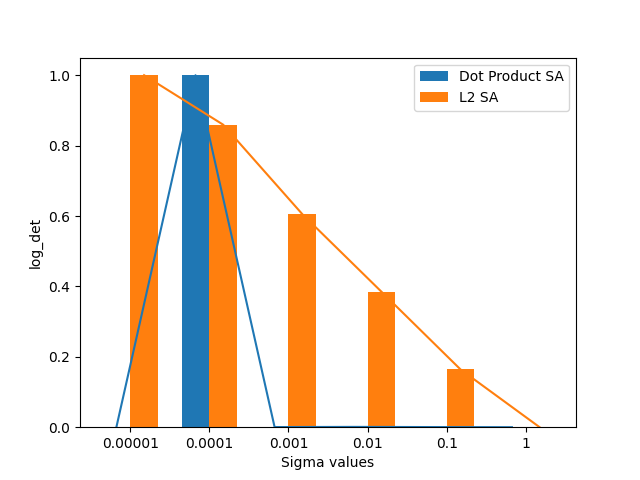}
      \hspace{-4.1em}
     \end{tabular}
  %   \vspace{-1.5em}
      \caption{Variation of (a)bits/dim, (b)log-likelihood and (c) volume correction values in ACF with dot-product and ACF with $L_2$ Self Attention with the change in intensity of Gaussian noises applied to perturb the input MNIST images. While the variations in the values with $L_2$ Self Attention are always consistent, the dot-product plots again fail to produce such consistent variations.}
      \label{fig:plots}
      \vspace{-1em}
   \end{figure*}

We perform ablations to study the robustness of Attentive Contractive Flows and validate the need for $L_2$ Self Attention over dot-product Self Attention by qualitatively and quantitatively evaluating the reconstructions of perturbed input images. For this experiment, we consider ACF(iResNet) over the MNIST dataset.  The input images are perturbed by adding Gaussian noises of different variances($\sigma$) - $0.00001, 0.0001, 0.001, 0.01, 0.1, 1$. We quantitatively evaluate the model by reporting the bits/dim values of the reconstructions for the different levels of noise. We compare the performance of  $L_2$  Self-Attention-based ACF with dot product Self-Attention-based ACF in Table \ref{tab:perturb}. The fact that dot-product self-attention is not Lipschitz-constrained leads to wrongful computation of the log-probability using the change of variable formula, which sometimes results in  negative bits/dim, as we can see in Table \ref{tab:perturb}.On the other hand, we observe that the model with $L_2$ Self Attention performs better and more consistently. 
The qualitative results in Figure \ref{recons_pert_$L_2$} also demonstrate that the reconstructions for ACF($L_2$) improve with decreasing values of $\sigma$. However, such is not the case with dot-product Self Attention. The robustness of using $L_2$ Self Attention with ACF is also graphically demonstrated in Figure \ref{fig:plots}(a),(b) and (c). While ACF with $L_2$ Self Attention follows the natural and consistent increase in bits/dim, dot product Self Attention fails to perform as expected and produces random undesirable values. We notice similar behavioural consistencies w.r.t the change in the trace values and also the log probability values when $L_2$ SA is used instead of dot-product SA. 

\begin{table}[ht]
  % \vspace{-0.7em}
  
  \centering
  \begin{tabular}{ c|cc }  \hline
       Noise ($\sigma$) & ACF(dot-product SA) & ACF($L_2$SA)\\
       \hline
       0.1 & -726.63 & 42889.64 \\
       0.01 & 35772.76 & 4959.88 \\
       0.001 & -9298.64 & 512.88\\
       0.0001 & -15126.88 & 47.50\\\hline\\
  \end{tabular}
  
  \caption{Comparison of bits/dim of reconstructions of the perturbed input images with Gaussian noises with the corresponding variances on ACF with dot-product Self Attention and ACF with $L_2$ Self Attention.}
  \label{tab:perturb}
  % \vspace{-2em}
  \end{table}

\section{Qualitative Examples}
\label{sample}
\begin{figure*}[ht]
\centering
   (a) CIFAR10 \\
   \includegraphics[width=\linewidth]{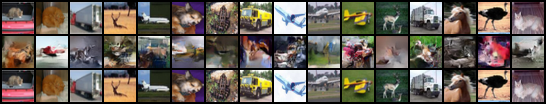}\\
   (b) ImageNet32  \\
   \includegraphics[width=\linewidth]{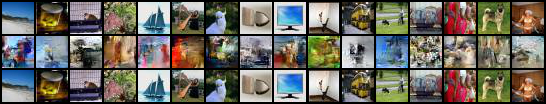}\\
   (c) ImageNet64 \\
   \includegraphics[width=\linewidth]{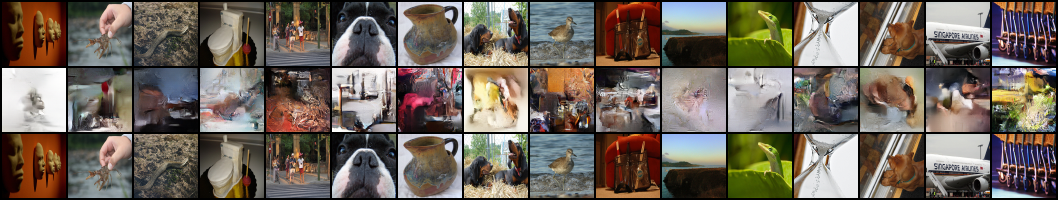} \\
   (d) 5-bit CelebA HQ \\
   \includegraphics[width=\linewidth]{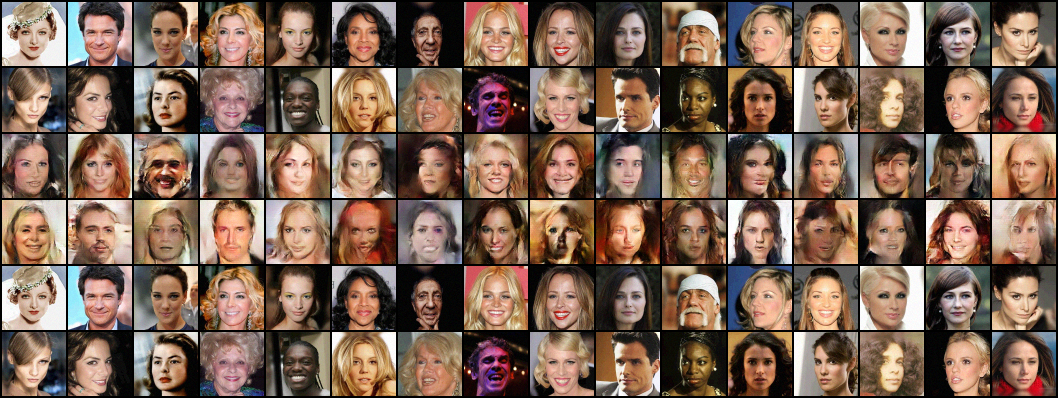}
 \vspace{1em}
\caption{(a,b,c) The images in the top, bottom and middle row are respectively the real, reconstructed and generated images from ACF (Residual Flow + $L_2$SA). (d) The top two rows represent the read images, the bottom two rows represent their reconstruction and the middle two rows are the generated samples from ACF (Residual Flow + $L_2$SA).}
\label{supp:samples_comparison}
% \vspace{-2em}
\end{figure*}

\begin{figure*}[h]
\centering
(a) CIFAR10 \\
\includegraphics[width=0.7\linewidth]{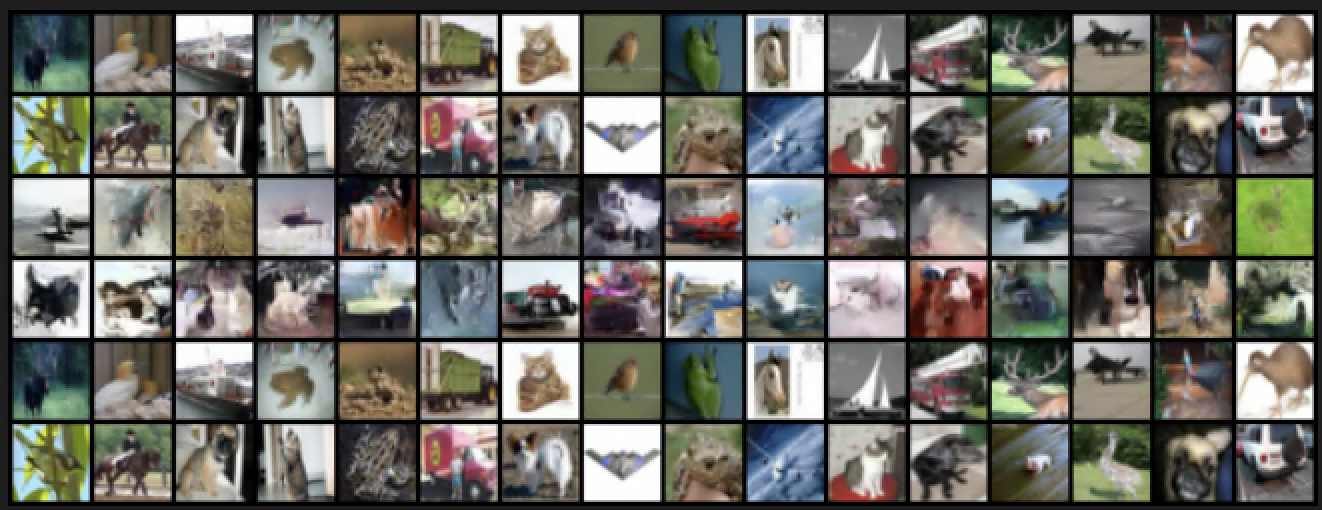} \\
(b) ImageNet32 \\
\includegraphics[width=0.7\linewidth]{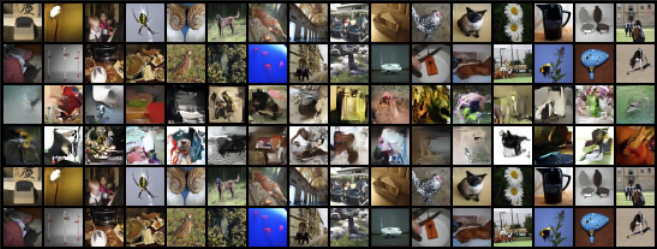} \\
\vspace{1em}
\caption{
Examples on CIFAR10 (above) and ImageNet32 (below) dataset from ACF(iDenseNet + $L_2$SA). For each dataset, the top two rows represent the read images, the bottom two rows represent their reconstruction and the middle two rows are the generated samples from the model.}
%Algorithm \ref{interpolate_algo}.}
\label{supp:fig:samples_iD}
\end{figure*}

\begin{figure}[ht]
  \centering
  \subfigure[MNIST]{%
    \includegraphics[width=0.45\linewidth]{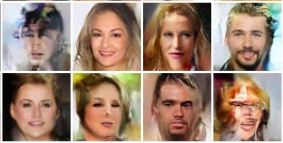}
  }
  % \hfill
  \subfigure[CIFAR10]{%
    \includegraphics[width=0.45\linewidth]{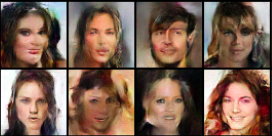}
  }
  \caption{Comparison between CelebA samples generated from vanilla residual flow and ACF (Residual Flow + $L_2$SA).}
  \label{supp:fig:compare}
\end{figure}

\begin{figure}[ht]
\centering
\includegraphics[width=0.7\linewidth]{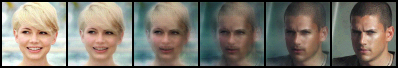}
\includegraphics[width=0.7\linewidth]{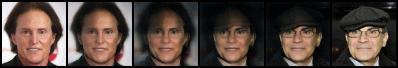}
\includegraphics[width=0.7\linewidth]{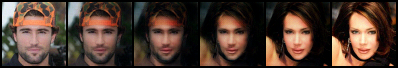}
\includegraphics[width=0.7\linewidth]{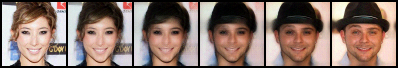}
\includegraphics[width=0.7\linewidth]{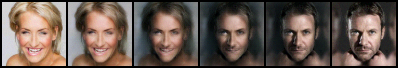}
\includegraphics[width=0.7\linewidth]{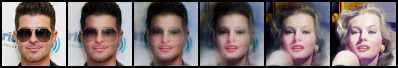}
\includegraphics[width=0.7\linewidth]{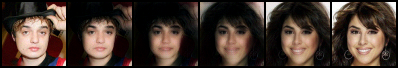}
\includegraphics[width=0.7\linewidth]{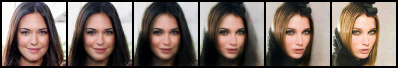}
\includegraphics[width=0.7\linewidth]{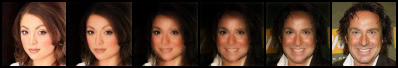}
\vspace{1em}
\caption{
Interpolation between CelebA images, from one face to another using ACF(Residual Flow + $L_2$SA)}
%Algorithm \ref{interpolate_algo}.}
\label{supp:fig:interpolations}
\end{figure}

Figure \ref{supp:samples_comparison} and \ref{supp:fig:samples_iD} show some more results from ACF models on various datasets. Figure \ref{supp:fig:compare} presents a qualitative comparison of the CelebA dataset between residual flow and ACF (Residual Flow). It can be visually observed that adding attention to the normalizing flow steps contributes in the generation of better samples. Interpolations between different CelebA face images while keeping the intermediate generations realistic are demonstrated in Figure \ref{supp:fig:interpolations}.

\newpage
% in documenet
% \begin{figure*}
%     \centering
%     \includegraphics[width=\linewidth]{Supp_Images/Samples/MNIST_ResFlow/ezgif.com-gif-maker.png}
%     \caption{Caption}
%     \label{fig:my_label}
% \end{figure*}
%\newpage
% \begin{figure*}[h]
% \centering
% (a) Residual Flow\\
% \includegraphics[width=0.7\linewidth]{Supp_Images/resflow-comp.png}\\
% (b) ACF (Residual Flow)\\
% \includegraphics[width=0.7\linewidth]{Supp_Images/acf-comp.png}\\
% \caption{
% Comparison between CelebA samples generated from vanilla residual flow and ACF (Residual Flow).}%Algorithm \ref{interpolate_algo}.}
% \label{fig:interpolations}
% \end{figure*}
% \begin{figure*}
%   \centering
%   \begin{subfigure}[b]{0.4\textwidth}
%     \centering
%     \includegraphics[width=\textwidth]{Supp_Images/resflow-comp.png}
%     \caption{Residual Flow}
%     \label{fig:y equals x}
%   \end{subfigure}
%   %\hfill
%   \begin{subfigure}[b]{0.4\textwidth}
%     \centering
%     \includegraphics[width=\textwidth]{Supp_Images/acf-comp.png}
%     \caption{ACF (Residual Flow)}
%     \label{fig:three sin x}
%   \end{subfigure}
%   \caption{Comparison between CelebA samples generated from vanilla residual flow and ACF (Residual Flow).}
%   \label{supp:fig:compare}
% \end{figure*}

\clearpage
\bibliography{BMVC/egbib}